\documentclass[times,1p]{elsarticle}
\usepackage{lineno}
\usepackage[colorlinks=true,linkcolor=black, citecolor=blue, urlcolor=blue]{hyperref}
\usepackage{url}

\usepackage{siunitx}
\usepackage{float}
\usepackage{subcaption}
\usepackage{tikz}
\usepackage{pgfplots}
\usepackage[english]{babel}
\usepackage{amsmath}
\usepackage{lipsum}  
\usepackage{dsfont}
\usepackage{graphicx}
\usepackage{wrapfig}
\usepackage[draft]{changes} 
\usepackage{hyperref}
\definechangesauthor[name={Reviewer 1}, color=orange]{R1.1}
\definechangesauthor[name={Reviewer 2}, color=orange]{R1.2}
\definechangesauthor[name={Reviewer 3}, color=orange]{R1.3} 

\DeclareSIUnit{\byte}{B}

\usetikzlibrary{positioning,calc,shapes,arrows,shapes.multipart}
\usepackage{enumitem}

\definecolor{color1bg}{HTML}{1f77b4}
\definecolor{color2bg}{HTML}{ff7f0e}
\definecolor{color3bg}{HTML}{2ca02c}
\definecolor{color4bg}{HTML}{d62728}

\tikzset{
	papDecision/.style = {
		diamond,
		fill = color3bg,
		aspect=2,
		draw, 
		text width = 20 mm, 
		align = center, 
		text badly centered,
		inner sep = 1 pt,
		font=\ttfamily\footnotesize,
		minimum width = 30mm,
		minimum height = 7mm,
	},
	papStart/.style = {
		rectangle,
		fill = color1bg,
		draw, 
		align = center, 
		text width = 3cm, 
		text badly centered,
		inner sep = 4 pt,
		rounded corners=10pt,
		font=\ttfamily\footnotesize,
		minimum width = 30mm,
		minimum height = 7mm,
	},
	papEnd/.style = {
		rectangle,
		fill = color1bg,
		draw, 
		align = center, 
		text width = 3cm, 
		text badly centered,
		inner sep = 4 pt,
		rounded corners=10pt,
		font=\ttfamily\footnotesize,
		minimum width = 30mm,
		minimum height = 7mm,
	},
	papData/.style = {
		trapezium,
		draw, 
		align = center, 
		text width = 20 mm, 
		text badly centered,
		inner sep = 4 pt,
		trapezium left angle=70,
		trapezium right angle=110,
		font=\ttfamily\footnotesize,
		minimum width = 30mm,
		minimum height = 7mm,
	},
	papPredProc/.style = {
		draw,
		rectangle split,
		rectangle split horizontal,
		rectangle split parts = 3,
		rectangle split empty part width=-8pt,
		align = center, 
		text badly centered,
		font=\ttfamily\footnotesize,
		minimum width = 30mm,
		minimum height = 7mm,
	},
	papProcess/.style = {
		rectangle,
		fill = color2bg,
		draw,
		align = center, 
		text width = 3cm, 
		text badly centered,
		font=\ttfamily\footnotesize,
		minimum width = 30mm,
		minimum height = 7mm,
	},
	papLine/.style = {
		draw,
		-stealth,
		font=\ttfamily\footnotesize,
	},
}

\begin{document}
\title{Scalable heliostat surface predictions from focal spots: Sim-to-Real transfer of inverse Deep Learning Raytracing}
\author[1,2]{Jan Lewen\corref{cor1}}
\ead{jan.lewen@dlr.de}
\author[1]{Max Pargmann}
\author[2]{Mehdi Cherti}
\author[2]{Jenia Jitsev}
\author[1]{Robert Pitz-Paal}
\author[1]{Daniel Maldonado Quinto}
\cortext[cor1]{Corresponding author}
\address[1]{German Aerospace Center (DLR), Institute of Solar Research, Linder Höhe, D-51147 Köln, Germany}
\address[2]{Research Center Jülich, Jülich Supercomputing Centre, Wilhelm-Johnen-Straße, 52428 Jülich, Germany}

\begin{abstract}
Concentrating Solar Power (CSP) plants are a key technology in the transition toward sustainable energy. A critical factor for their safe and efficient operation is the accurate distribution of concentrated solar flux on the receiver. However, flux distributions from individual heliostats are highly sensitive to surface imperfections, such as canting and mirror deformations. Measuring these surfaces across hundreds or thousands of heliostats remains impractical in real-world deployments. As a result, control systems often assume idealized heliostat surfaces, leading to suboptimal performance and potential safety risks. To address this, inverse Deep Learning Raytracing (iDLR) \cite{Lewen2025} has been introduced as a novel method for inferring heliostat surface profiles from target images recorded during standard calibration procedures. In this work, we present the first successful Sim-to-Real transfer of iDLR, enabling accurate surface predictions directly from real-world target images. We evaluate our method on 63 heliostats under real operational conditions. iDLR surface predictions achieve a median mean absolute error (MAE) of 0.17~mm and show good agreement with deflectometry ground truth in 84\% of cases. When used in raytracing simulations, it enables flux density predictions with a mean accuracy of 90\% compared to deflectometry over our dataset, and outperforms the commonly used ideal heliostat surface assumption by 26 \%. We tested this approach in a challenging double-extrapolation scenario—involving unseen sun positions and receiver projections—and found that iDLR maintains high predictive accuracy, highlighting its generalization capabilities. Our results demonstrate that iDLR is a scalable, automated, and cost-effective solution for integrating realistic heliostat surface models into digital twins. This opens the door to improved flux control, more precise performance modeling, and ultimately, enhanced efficiency and safety in future CSP plants.
\end{abstract}


\maketitle
\newlist{abbrv}{itemize}{1}
\setlist[abbrv,1]{label=,labelwidth=1in,align=parleft,itemsep=0.1\baselineskip,leftmargin=!}

\section*{List of Abbreviations}
\sectionmark{List of Abbreviations}

\begin{abbrv}
\item[CSP]          Concentrating Solar Power
\item[STJ]          Solar Tower Jülich
\item[DLR]          Deutsches Zentrum für Luft- und Raumfahrt
\item[iDLR]			inverse Deep Learning Raytracing
\item[GAN]          Generative Adversarial Network
\item[GPU]          Graphics Processing Unit
\item[NURBS]        Non-Uniform Rational B-Spline 
\item[MAE]          Mean Absolute Error
\item[SSIM]         Structural Similarity Index
\item[CSR]          Circumsolar Ratio
\end{abbrv}
    
\section{Introduction} \noindent
Concentrated Solar Power (CSP) is a key technology for the decarbonization of energy systems, offering the capability to deliver dispatchable electrical energy~\cite{Schniger2021} and to drive solar thermochemical reactions for fuel production~\cite{Schppi2021}. A crucial aspect of efficient CSP plant operation is the flux density distribution on the receiver, as each receiver has specific thermal thresholds. Proper flux distribution minimizes material stress and degradation, thereby optimizing receiver performance. While the effects of tracking errors and the hence mandatory automatized calibration have been extensively studied~\cite{stone1986targetmethod, Sun2015, Sattler2020-st, Sarr2021, Dring2022, Pargmann2023}, recently obtaining accurate flux density predictions for individual heliostats from easily available data sources for further optimized receiver performance has become a focus of research~\cite{Zhu2022, Pargmann2024, Kuhl2024_flux_pred, Lewen2025}. \newline \newline Traditionally, highly accurate flux density predictions required reconstructing the heliostat surface, accounting for mirror and canting errors, and simulating the resulting flux using raytracing techniques~\cite{Ulmer2011-gp, Belhomme2009}. Established methods for measuring heliostat surface deformation include deflectometry~\cite{Ulmer2011-gp}, laser scanning~\cite{photogrammetry_and_laserscanning, monterreal2017improved}, photogrammetry~\cite{Shortis1996, Pottler2005-yb, pottler2005photogrammetry, roger2010heliostat, bonanos2019heliostat}, and flux mapping~\cite{MARTINEZHERNANDEZ2023112162} (see \citet{ARANCIBIABULNES2017673_Review} for a review). Despite their accuracy, these methods require manual labor and additional hardware, making them expensive and impractical for commercial-scale deployment aimed at creating digital twins of heliostats for better heliostat control. Hence, in most cases flux density predictions for heliostat control are made either through raytracing of an ideal heliostat or by adopting simplified flux density assumptions. To overcome this modelling error undermining the safe and efficient operation of the power plant, several works were done to obtain very precise flux density predictions without the need of new hardware installation and manual work. \newline \newline \citet{Zhu2022} introduced a post-installation calibration procedure that estimates four geometric parameters per facet through optimization, using only target images of the heliostat's focal spot. Despite the simplicity of the heliostat model—assuming only canting and focusing errors—the approach demonstrated significant efficiency gains up to 20\%. This highlights the potential of integrating accurate flux density prediction into aim point optimization strategies. However, the model used by \citet{Zhu2022} is limited in scope. For example, the heliostats at Solar Tower Jülich (STJ) in Germany have 4 canted facets, however the individually facets are flat and non-focussing. Their flux density is significantly influenced by surface characteristics such as waviness and facet edge bending, which are not captured by their method. To address this, \citet{Pargmann2024} proposed a more generic heliostat model based on a differentiable raytracing framework. Their method optimizes a NURBS spline to match target images, allowing for more accurate flux density prediction. Nevertheless, due to the ill-posed and underdetermined nature of the inverse problem, the surface prediction is in most cases poor. This introduces uncertainty when extrapolating flux density from the lambertian target plane to the receiver plane with potentially complex geometry. Alternatively, \citet{Kuhl2024_flux_pred} presented a fully data-driven approach, representing each heliostat by a latent vector with no physical interpretation. This method achieves highly accurate flux density predictions and is straightforward to implement, as it requires no physical modeling. However, the lack of interpretability limits its integration with other physically grounded factors affecting flux density, such as soiling models, sun shape effects, shading, blocking, or extrapolation to complex receivers. To overcome the limitations of the above mentioned approaches, \citet{Lewen2025} recently introduced inverse Deep Learning Raytracing (iDLR), a novel deep learning approach that infers heliostat surface information from target images of focal spots. This method leverages the shape of the flux distribution to predict the underlying surface characteristics. iDLR has demonstrated high accuracy and strong agreement with state-of-the-art surface measurement techniques such as deflectometry, based on simulated flux densities. However, up to date, its application was only shown on synthetic data, and the model has not yet been transfered to real-world measurements.
\newline \newline 
Recent advancements in deep learning have been fueled by the availability of large-scale datasets. However, across domains such as robotics \cite{Tobin2017, Tobin2017DomainRA, Peng2018}, autonomous driving \cite{Yue2019, Pouyanfar2019}, and drone navigation \cite{Drone_domain_rand} or as mentioned above iDLR \cite{Lewen2025} exclusive reliance on real-world data for training is often infeasible due to the high costs and time required for data acquisition. Consequently, simulated data play a central role in training models that are ultimately deployed in real-world settings. However, the inherent domain discrepancy between synthetic and real data—commonly referred to as the sim-to-real gap \cite{Tobin2017}—poses a significant challenge to direct transferability, often leading to poor generalization in real-world environments. To enable a robust Sim-to-Real transfer of iDLR, three general strategies are conceivable. The first is fine-tuning the model on a limited amount of real-world data \cite{VOOGD20231510}. The second involves \textit{Domain Adaptation}, where the discrepancy between the simulated and target domains is bridged, often using unsupervised deep learning models such as Generative Adversarial Networks (GANs) \cite{hoffman2017cycadacycleconsistentadversarialdomain, tzeng2017adversarialdiscriminativedomainadaptation, Bewley, Hong2018}. The third strategy is \textit{Domain Randomization}, which aims to inject sufficient variability into the simulation environment such that the distribution of real-world data becomes a subset of the expanded simulated domain \cite{Tobin2017, tobin2018domainrandomizationgenerativemodels, Peng2018}. Among these, fine-tuning appears to be the most straightforward. However, this approach presents significant drawbacks. It requires paired real deflectometry measurements and target images, which are still expensive and challenging to obtain. More importantly, this would necessitate the deployment of surface measurement infrastructure at every power plant where iDLR is to be used—undermining the very advantage of eliminating the need for additional costly hardware. Therefore, this approach is impractical and counterproductive for broader adoption in CSP systems, and we exclude it in favor of a zero-shot Sim-to-Real strategy where no further real target images are required for training. \newline \newline 
In this work, we propose a zero-shot Sim-to-Real transfer method for inverse Deep Learning Raytracing (iDLR) that enables accurate heliostat surface reconstruction directly from images of their focal spots. Our approach combines Domain Adaptation and Domain Randomization to bridge the sim-to-real gap without training on difficult obtainable pairs of real world data. For Domain Adaptation, we follow the strategy of \citet{Zhang2019}, transforming real-world data to resemble simulated data instead of the other way around. Specifically, we leverage the deep learning framework by \citet{Kuhl2024}, which employs a UNet architecture trained exclusively on raytraced flux distributions to predict underlying irradiance maps from target images. As a result, real-world observations are implicitly projected into the simulated raytracing domain. To further enhance generalization, we apply Domain Randomization during training, exposing the model to a wide range of variations, that resemble the difference between real target images and simulated data. Together, these techniques enable zero-shot generalization of iDLR to real data, making the approach highly practical and keeping it cost-efficient. We validate our method across 63 heliostats under operational conditions and demonstrate strong agreement with deflectometry-based ground truth for the far majority of heliostats (84\%), achieving a median surface error of 0.17~mm and high accuracy in flux prediction. Even under challenging extrapolation scenarios—e.g., predicting fluxes on the receiver under unseen sun positions—iDLR maintains robust performance. These results establish iDLR as a scalable and physically interpretable component for heliostat digital twins, with significant potential for improving monitoring, calibration, and control in future CSP plants.

\section{Method}\noindent
In this section, we present the steps for transferring iDLR to real world data. 
\begin{figure}[h]
    \includegraphics[width=\columnwidth]{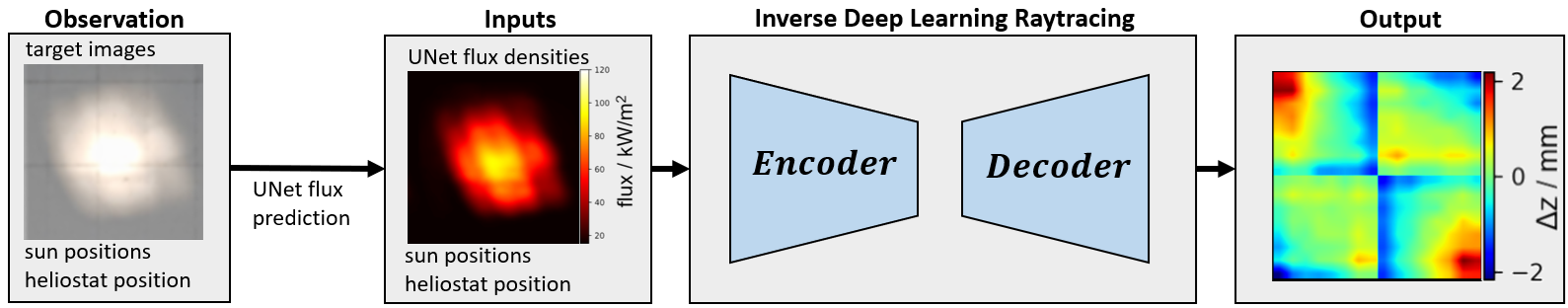}
    \caption{Overview of the full inverse Deep Learning Raytracing (iDLR) pipeline. Target images are first processed by the UNet model proposed by \citet{Kuhl2024}, which translates them into flux density predictions. Up to eight such predictions are then input to the iDLR model, which has been trained solely on simulated data using Domain Randomization. The iDLR model subsequently infers the heliostat surface shape from these flux density inputs.}
    \label{fig:iDLR_pipeline}
\end{figure}\noindent 
The general workflow of the iDLR approach is illustrated in Figure~\ref{fig:iDLR_pipeline}. For each heliostat, a series of target images capturing the flux spot are acquired as part of the standard calibration procedure. These images are processed using the UNet-based method proposed by \citet{Kuhl2024}, which predicts flux density distributions—hereafter referred to as UNet flux density. For each heliostat, up to eight such UNet flux densities, along with the corresponding sun positions and the heliostat’s position, are provided as input to the iDLR model. The iDLR model, which is trained on a semi-artificial dataset constituted by deflectometry surfaces and simulated flux densities, then predicts the underlying surface shape of the heliostat. In the following we will describe the data needed for training the model, the model architecture and the training routine.

\subsection{Data}
\begin{figure}[h]
    \includegraphics[width=\columnwidth]{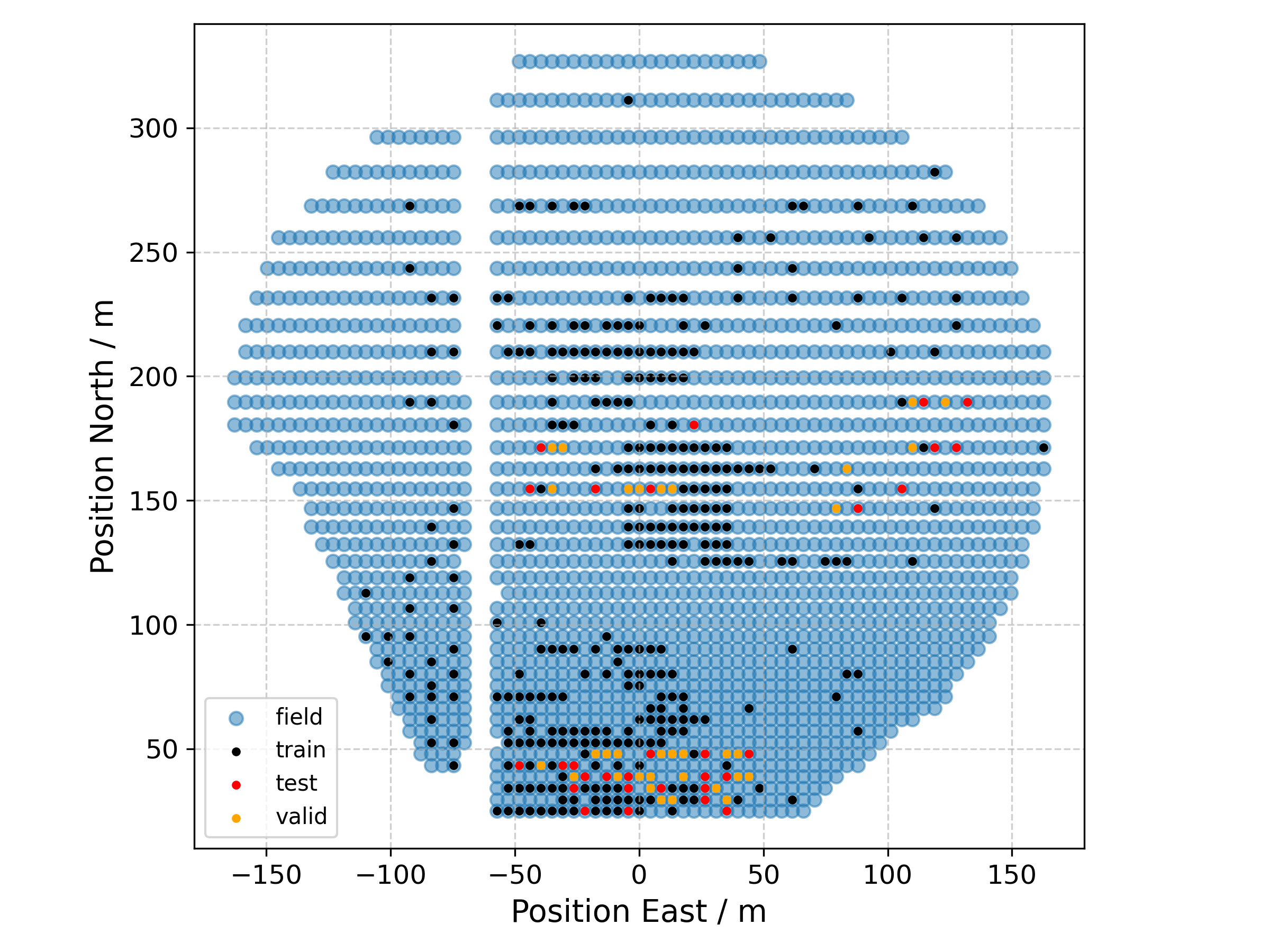}
    \caption{The heliostat field of the Solar tower Jülich. The heliostat's which were used for training, validating and testing the iDLR model are highlited.}
    \label{fig:heliostat_field}
\end{figure}

\noindent
The key real data for training an iDLR model are measurements of heliostat surfaces. Since 2014, more than 600 deflectometry measurements have been conducted at the STJ facility in Germany (see \citet{2025paint} for data collection from STJ) \footnote{Link to data collection: \url{https://paint-database.org}}. Throughout this period, the same heliostat model was used—comprising four canted, flat non-focussed facets, each measuring 1.6m$\times$1.25m. The focal distance of the canting is for every heliostat the distance to the main receiver, or to a side receiver, whereby the minimum focal distance is 65~m. Between 2020 and 2022, modifications were made to the heliostat kinematics. As a result, the surface error characteristics of the heliostats changed, altering the corresponding flux density distributions. Consequently, older deflectometry measurements could not be used for validation or testing, as the flux patterns no longer match to the old heliostat surface. Nevertheless, because our model is trained on simulated flux densities via raytracing, these earlier measurements remain suitable for training. To ensure data quality, deflectometry measurements with poor coverage or visible artifacts were excluded. A measurement was only considered valid if at least 95\% of the heliostat surface was successfully captured. Additionally, visual inspection was performed to identify and discard measurements with potential artifacts. After this filtering process, a total of 456 deflectometry measurements were considered valid and included in our approach. \newline \newline Each heliostat surface is represented using the NURBS-based model introduced by \citet{Lewen2025}. Specifically, the surface of each facet is parametrized by an $8 \times 8$ grid of z-control points $P_{i,j}^{z}$ that define the corresponding NURBS spline. The dataset was split into training, validation, and testing subsets. A total of 393 measurements were used for training, representing 323 unique heliostats (as some were measured multiple times). The validation and test sets consist of 33 and 30 individual heliostats, respectively. All heliostats in the validation and test dataset are canted on the main receiver. Their positions within the STJ heliostat field are illustrated in Figure~\ref{fig:heliostat_field}. \newline \newline As training a deep learning model on only 393 data points is difficult, augmentation techniques were applied to create new, but still realistic, synthetic heliostat surfaces. To augment the training dataset, two surface augmentation techniques were employed. First, heliostat surfaces were rotated by 180~$\deg$. Second, new surfaces were generated by computing weighted averages between pairs of existing surfaces. In total, approximately 400{,}000 augmented heliostat surfaces were created for model training.
\noindent
In the next step, flux density distributions were simulated using a raytracer. Compared to the approach described in \citet{Lewen2025}, our procedure differs in four aspects: First, in \citet{Lewen2025}, sun positions were sampled based on those at which real target images were taken. This approach introduced a bias, as summer sun positions were overrepresented. To mitigate this, we now sample sun positions from an equidistant grid in azimuth-elevation space, covering all possible sun positions throughout the year at the geographic location of the STJ. Second, we revised the normalization strategy for the flux densities. Previously, the sum of all pixel values in a flux density map was normalized to 100, effectively preserving total energy. However, real target images do not conserve energy in this way. Thus, we now normalize each flux density such that its maximum value is 1, which better reflects the characteristics of real measurements. Third, in line with the Domain Randomization strategy, we uniformly sample the Circumsolar Ratio (CSR) parameter for the Buie-CSR sunshape model within the range of 0 to 0.15. Fourth, the heliostat aim point is randomly scattered within a 1~m radius around the target center. These two randomized parameters are integrated directly into the simulation process to improve model robustness. Additional randomization steps are applied during training to further diversify the dataset. \newline \newline While training is performed on this semi-artificial dataset, validation and testing are conducted using real-world target images. At the STJ, several target images are acquired annually for calibration purposes. The number of images per heliostat varies depending on the already achieved quality of the calibration. The target images were filtered to remove those affected by artifacts such as overexposure, overlapping flux spots from multiple heliostats, or shading and blocking. The final dataset includes between three and eight target images per heliostat. Then, in accordance with the Domain Adaptation strategy for the Sim-to-Real transfer, the UNet model introduced by \citet{Kuhl2024} is applied to these target images to generate flux density predictions from the target image, which serve as input for inference with the simulatively trained iDLR model. The validation dataset is used during training to monitor model performance and guide architectural and hyperparameter decisions based on validation loss. The test dataset is used solely to evaluate the final model after training. As no significant performance differences were observed between the validation and test sets, results presented in the subsequent section are reported jointly for both datasets to improve statistical robustness.
\subsection{Model} 
\begin{figure}[h]
    \includegraphics[width=\columnwidth]{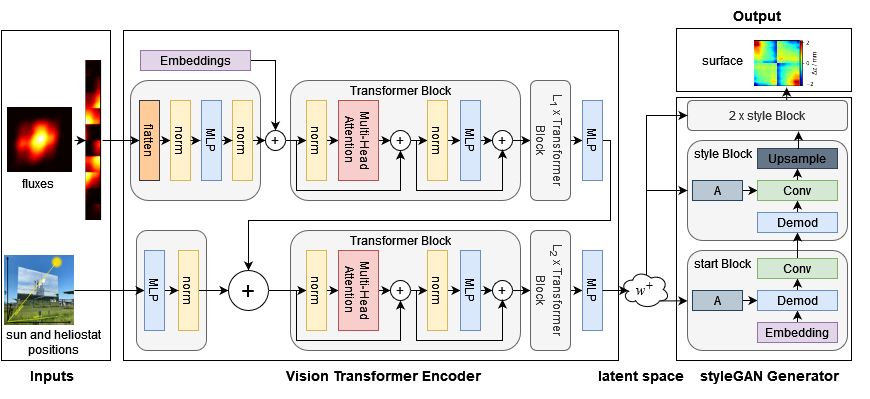}
    \caption{The iDLR architecture consists of a Vision Transformer based Encoder and a styleGAN2 Generator.}
    \label{fig:model}
\end{figure}

\noindent
The model architecture is illustrated in Figure~\ref{fig:model}. It follows an encoder-decoder design, where the encoder consists of two primary components: a Vision Transformer (upper path), as introduced by \citet{dosovitskiy2021imageworth16x16words}, and a modified standard Transformer Encoder (lower path), originally proposed by \citet{vaswani2023attentionneed}. The Vision Transformer is responsible for processing the input flux density distributions. First, each flux density map is divided into patches of size 16, which are then passed through a Multi-Layer Perceptron (MLP). Positional embeddings and a class embedding are added to the patch embeddings before they are input into the Transformer Block. The core computation occurs in the Multi-Head Attention Block, where four parallel attention heads are applied. The output is then passed through another MLP, and a residual connection is added. This Transformer Block is repeated $L_1 = 8$ times, and all MLP layers within this module have a dimensionality of 128. The output is a compact feature vector of dimension 128. This feature vector is then combined with positional information describing the heliostat and sun positions. These positional inputs are processed by an MLP to produce embeddings, which are then summed with the flux density feature vector. The combined representation is passed into the second part of the encoder—a modified Transformer Encoder. This module uses Multi-Head Attention with five attention heads and is repeated $L_2 = 8$ times. The output of this encoder is a latent representation encoding the heliostat surface geometry. The decoder is implemented using a StyleGAN2 generator \cite{karras2019stylebased, karras2020analyzing}, which operates in the w+w+ latent space as introduced by \citet{Richardson2021}. Since the generator comprises three style blocks, the w+ latent vector has a dimensionality of $3 \times 32$, with the value 32 selected empirically. Each latent vector is projected via an affine transformation and weight demodulation into the generator, which includes a learnable constant input in the initial block, followed by convolutional and upscaling layers. To ensure stable training and consistent feature scaling, layer normalization is applied at the locations indicated in Figure~\ref{fig:model}. Additionally, although not shown in the figure, a Dropout layer with a dropout rate of 0.2 is applied after each MLP to mitigate overfitting. In total the model has slightly more than 4 million trainable parameter.
\noindent 

\subsection{Training}\noindent
The training procedure differs in one key aspect from that described in \citet{Lewen2025}: the incorporation of randomization techniques during training to improve the model's robustness when applied to real-world data. Although two forms of randomization were already introduced during the simulation process, we chose—for performance and flexibility reasons—to primarily apply additional randomizations directly to the precomputed flux densities during training. All modifications to the flux densities are designed to reflect plausible physical phenomena. Discrepancies between raytraced flux density simulations and UNet-predicted images typically arise from two major sources: (1) latent parameters related to the heliostat and the environmental conditions at the power plant, and (2) measurement errors or artifacts introduced by the camera system. Latent parameters include, for example, variations in the sunshape, mirror soiling, shading and blocking effects, microscopic surface errors, inaccuracies in coordinate data, non-ideal Lambertian target behavior, and background radiation. Camera-related artifacts include overexposure, underexposure, and uncertainties in contrast settings. To improve the model's ability to generalize to real-world measurements, the following randomizations were introduced during training:
\begin{itemize} \item Random exclusion of individual target images, encouraging the model to perform well with fewer observations. \item Uniform clamping of flux densities in the range (0.9, 1.0) to simulate overexposure effects. \item Small perturbations to the heliostat and sun positions to simulate coordinate uncertainties. \item Additive noise to the surface representation to mimic potential errors in the deflectometry data. \item Background noise distributed over the flux density map to represent environmental radiation. \item Random adjustment of contrast, simulating the unknown contrast parameter of the camera system. \item Random cropping by one pixel on a randomly selected subset of image edges (each edge cropped at most once), followed by interpolation back to the original resolution. This transformation effectively alters the size and shape of the flux density and can mimic effects such as soiling, microscopic mirror deformation, and inaccuracies in the Lambertian target dimensions. \item Small random deformations applied to the flux density, primarily to simulate deviations from ideal Lambertian reflection behavior. \item Bilinear smoothing to slightly blur and broaden the flux distribution, diminishing fine-grained details. \end{itemize}
Each of the above randomizations is applied independently with a probability of 0.5 per flux density in each training batch. As a result, the model almost never encounters an unaltered image during training, promoting strong generalization and robustness. \newline \newline 
The model was trained for 50 epochs using four NVIDIA A100 GPUs, with a total training time of approximately five hours, including the computational overhead of validation and testing via raytracing. The initial learning rate was set to 0.001 and was exponentially decayed using a gamma factor of 0.995. Weight decay regularization was applied with a factor of $1 \times 10^{-7}$. The Adam optimizer was employed with parameters $\beta = (0.9, 0.999)$ and $\varepsilon = 1 \times 10^{-8}$. The training objective is to minimize the \textit{Mean Absolute Error} (MAE) between the predicted z-control points, denoted as $P_{i,j}^{z, \text{iDLR}}$, and those derived from deflectometry measurements, $P_{i,j}^{z, \text{Defl}}$:
\begin{equation} \label{eq:MAE} \text{MAE}(P^{z, \text{iDLR}}, P^{z, \text{Defl}}) = \frac{1}{N} \sum_{i=1}^{n=8} \sum_{j=1}^{m=8} \left| P_{i,j}^{z, \text{iDLR}} - P_{i,j}^{z, \text{Defl}} \right| \end{equation}
Where: \begin{itemize} \item nn and mm denote the number of control points in the horizontal and vertical directions, respectively. \item N=n×mN=n×m is the total number of control points per facet. \end{itemize}
Although MAE is an intuitive and widely used error metric, it is sensitive to the absolute magnitude of surface deviations. Since individual heliostats can vary significantly in terms of canting and surface deformation, MAE alone may not provide a fully representative measure of prediction quality across heliostats with different error magnitudes. To address this, we also evaluate the \textit{Structural Similarity Index} (SSIM), a perceptual metric that better captures the structural similarity between the predicted and measured heliostat surfaces. SSIM is defined as:
\begin{equation} \label{eq:ssim} \text{SSIM}(P^{z, \text{iDLR}}, P^{z, \text{Defl}}) = \frac{(2\mu_{P^{z, \text{iDLR}}} \mu_{P^{z, \text{Defl}}} + C_1)(2\sigma_{P^{z, \text{iDLR}}, P^{z, \text{Defl}}} + C_2)}{(\mu_{P^{z, \text{iDLR}}}^2 + \mu_{P^{z, \text{Defl}}}^2 + C_1)(\sigma_{P^{z, \text{iDLR}}}^2 + \sigma_{P^{z, \text{Defl}}}^2 + C_2)} \end{equation}
Where:
\begin{itemize}
    \item \( P^{z, \text{iDLR}} \) and \( P^{z, \text{Defl}} \) are the iDLR and Deflectometry z-control points.
    \item \( \mu_{P^{z, \text{iDLR}}} \), \( \mu_{P^{z, \text{Defl}}} \) are the mean values of \( P^{z, \text{iDLR}} \) and \( P^{z, \text{Defl}} \).
    \item \( \sigma_{P^{z, \text{iDLR}}}^2 \), \( \sigma_{P^{z, \text{Defl}}}^2 \) are the variances of \( P^{z, \text{iDLR}} \) and \( P^{z, \text{Defl}} \).
    \item \( \sigma_{P^{z, \text{iDLR}}, P^{z, \text{Defl}}} \) is the covariance between \( P^{z, \text{iDLR}} \) and \( P^{z, \text{Defl}} \).
    \item \( C_1 \) and \( C_2 \) are small constants to stabilize the division when the denominators are close to zero.
\end{itemize}
SSIM values range from \(-1\) to \(1\), where higher values indicate greater structural similarity. The interpretation of SSIM values for surface comparison is as follows:
\begin{itemize}
    \item \(0.75 \leq \text{SSIM} \leq 1.0\): Very High similarity
    \item \(0.50 \leq \text{SSIM} < 0.75\): High similarity
    \item \(0.25 \leq \text{SSIM} < 0.50\): Medium similarity, usually the main features are still predicted good
    \item \(-0.25 < \text{SSIM} < 0.25\): No similarity
    \item \(\text{SSIM} \leq -0.25\): Negative similarity
\end{itemize}
SSIM is more robust to changes in absolute magnitude and focuses instead on the structural consistency of predicted surfaces, making it the preferred metric for comparing individual heliostats with varying deformation levels. However, for comparison of different models on the same dataset the MAE is preferably used because of its easier physical interpretability.

\section{Results}\noindent
\begin{figure}[h]
    \includegraphics[width=\columnwidth]{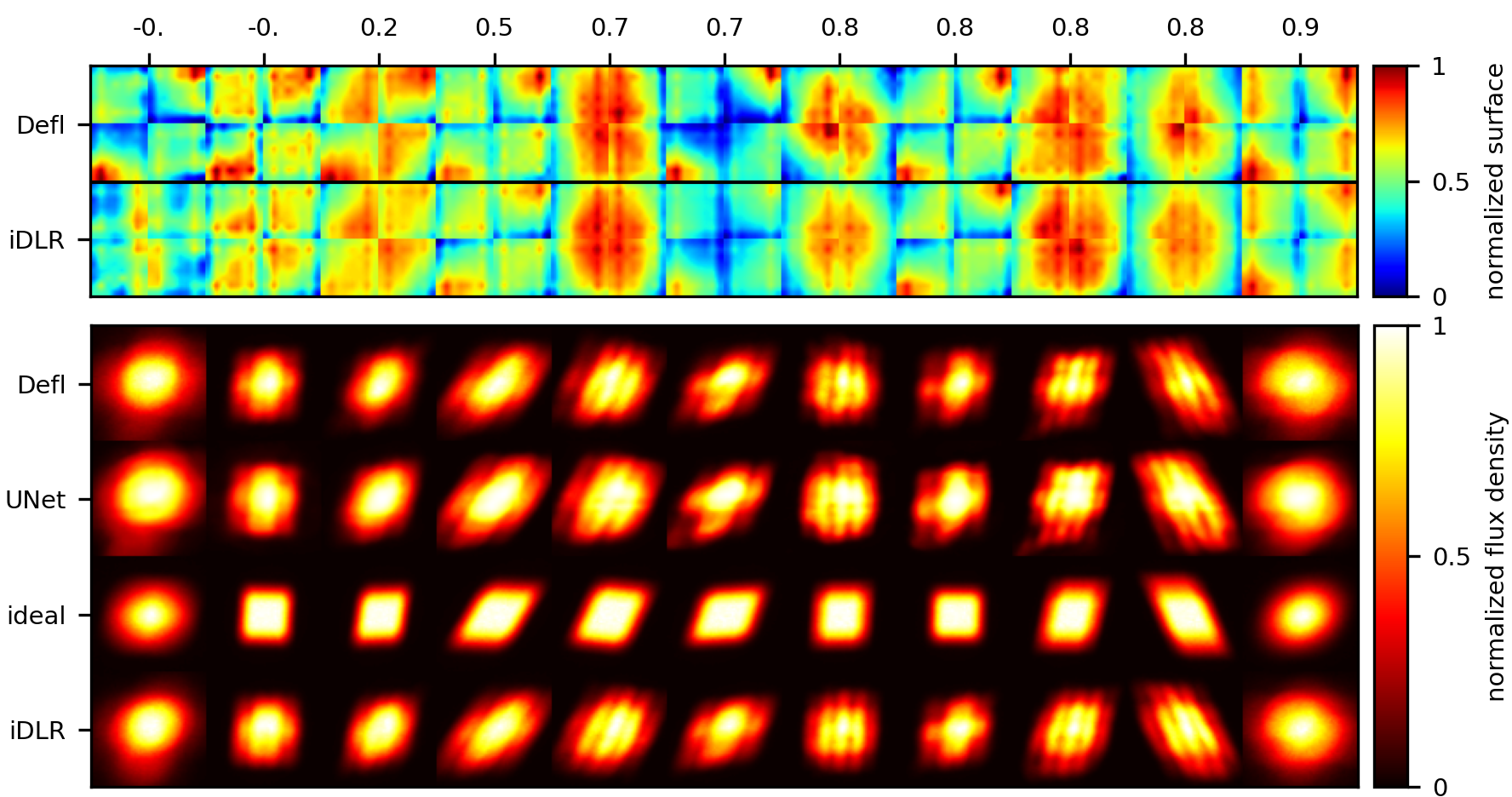}
    \caption{The top panel compares the iDLR surface prediction with Deflectometry measurements for ten heliostats at STJ. Above each prediction the SSIM between iDLR surface and Deflectometry is given. The lower panel compares the flux density from iDLR enhanced raytracing with Deflectometry enhanced raytracing, the UNet flux density prediction derived from a target image and the ideal heliostat assumption.}
    \label{fig:results_iDLR}
\end{figure}\noindent 
Figure~\ref{fig:results_iDLR} presents a visual comparison of the iDLR surface predictions against deflectometry measurements (top row), alongside a comparison of the resulting flux density distributions generated through iDLR-enhanced raytracing, deflectometry-enhanced raytracing, UNet flux densities, and an idealized heliostat assuming no surface errors. The surface predictions are ordered by their Structural Similarity Index (SSIM), ranging from the lowest (left) to the highest (right). The corresponding SSIM value is annotated above each surface prediction. For the 63 heliostats in the combined validation and test datasets, the MAE statistics (Minimum, First Quartile, Median, Mean, Third Quartile, Maximum) are (0.08, 0.14, 0.17, 0.18, 0.22, 0.34)~mm, while corresponding SSIM values are (-0.02, 0.54, 0.77, 0.66, 0.86, 0.94). These metrics indicate a high degree of accuracy in surface prediction for the majority of heliostats. However, a small subset of predictions exhibit significant deviations. Based on our observations, an SSIM value above 0.25 typically reflects the correct reconstruction of at least the main structural features of the surface. Therefore, we define a misprediction as any result with $SSIM<0.25$. According to this criterion, 16\% of the predictions are considered mispredictions, while 84\% exhibit similarity to the deflectometry ground truth. Flux density predictions obtained through iDLR-enhanced raytracing, when compared to deflectometry-enhanced raytracing under novel sun positions, achieve accuracy values (Min, Q1, Median, Mean, Q3, Max) of (0.80, 0.90, 0.92, 0.91, 0.93, 0.96). Notably, even for cases where surface prediction quality is relatively poor, the resulting flux density predictions usually remain close to the deflectometry-based ground truth. This can be attributed to the underdetermined nature of the inverse problem, where multiple surfaces may yield similar flux distributions. \newline \newline
\begin{figure}[h]
    \includegraphics[width=\columnwidth]{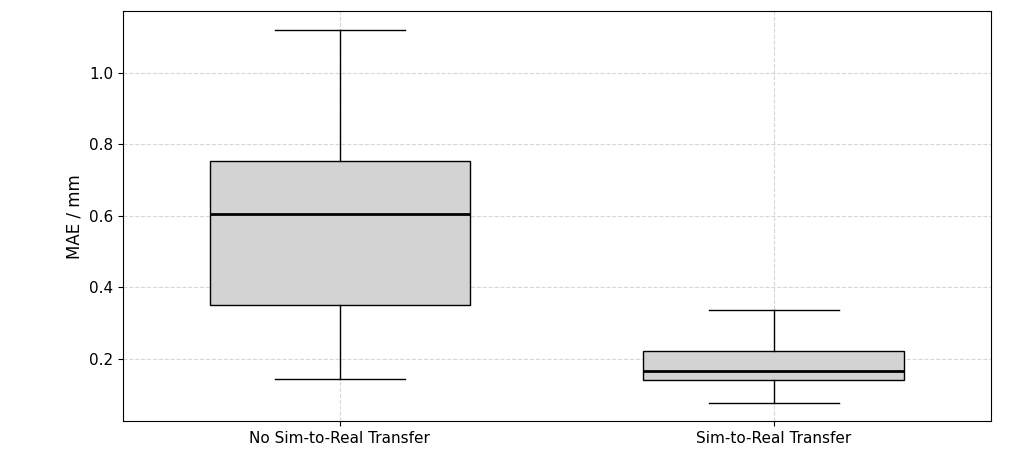}
    \caption{The left boxplot shows the model performance without Sim-to-Real techniques, while the right boxplot shows the significantly improved performance when the full Sim-to-Real pipeline is applied.}
    \label{fig:sim2real}
\end{figure}\noindent 
To assess the effectiveness of our Sim-to-Real transfer approach, we compare the surface prediction performance of the full iDLR pipeline against a baseline model without Sim-to-Real transfer. In the baseline setup, flux density inference is performed directly on raw target images, bypassing the UNet-based Domain Adaptation. Additionally, no Domain Randomization techniques are applied during training. Figure~\ref{fig:sim2real} presents the distribution of the MAE for both configurations. Without Sim-to-Real transfer, the model fails to generalize to real-world data in most cases, resulting in a more than threefold increase in the MAE. This highlights the critical role of Sim-to-Real transfer in enabling accurate heliostat surface reconstruction from real target images using iDLR.\newline \newline
\begin{figure}[h]
    \includegraphics[width=\columnwidth]{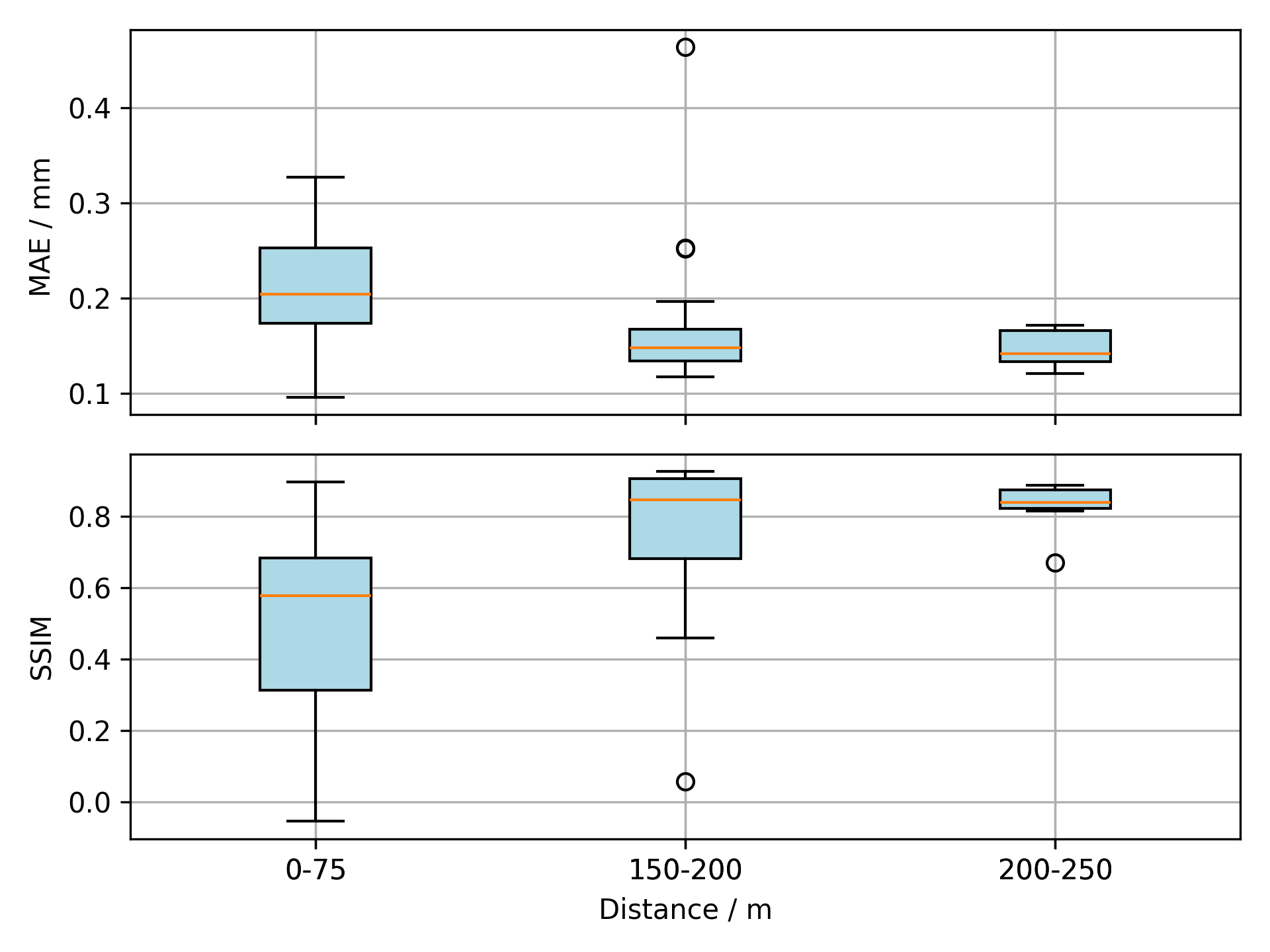}
    \caption{Surface prediction error over the distance of the heliostat to the tower.}
    \label{fig:surface_distance}
\end{figure}\noindent 
Figure~\ref{fig:surface_distance} illustrates the distribution of the surface prediction error as a function of the heliostat's distance to the tower, evaluated using both the MAE and SSIM. For both metrics, the accuracy of the surface predictions improves with increasing distance from the tower. Notably, not only does the average prediction error decrease, but the variance of the predictions also becomes smaller, with the exception of a few outliers. \newline \newline
\begin{figure}[h] 
\includegraphics[width=\columnwidth]{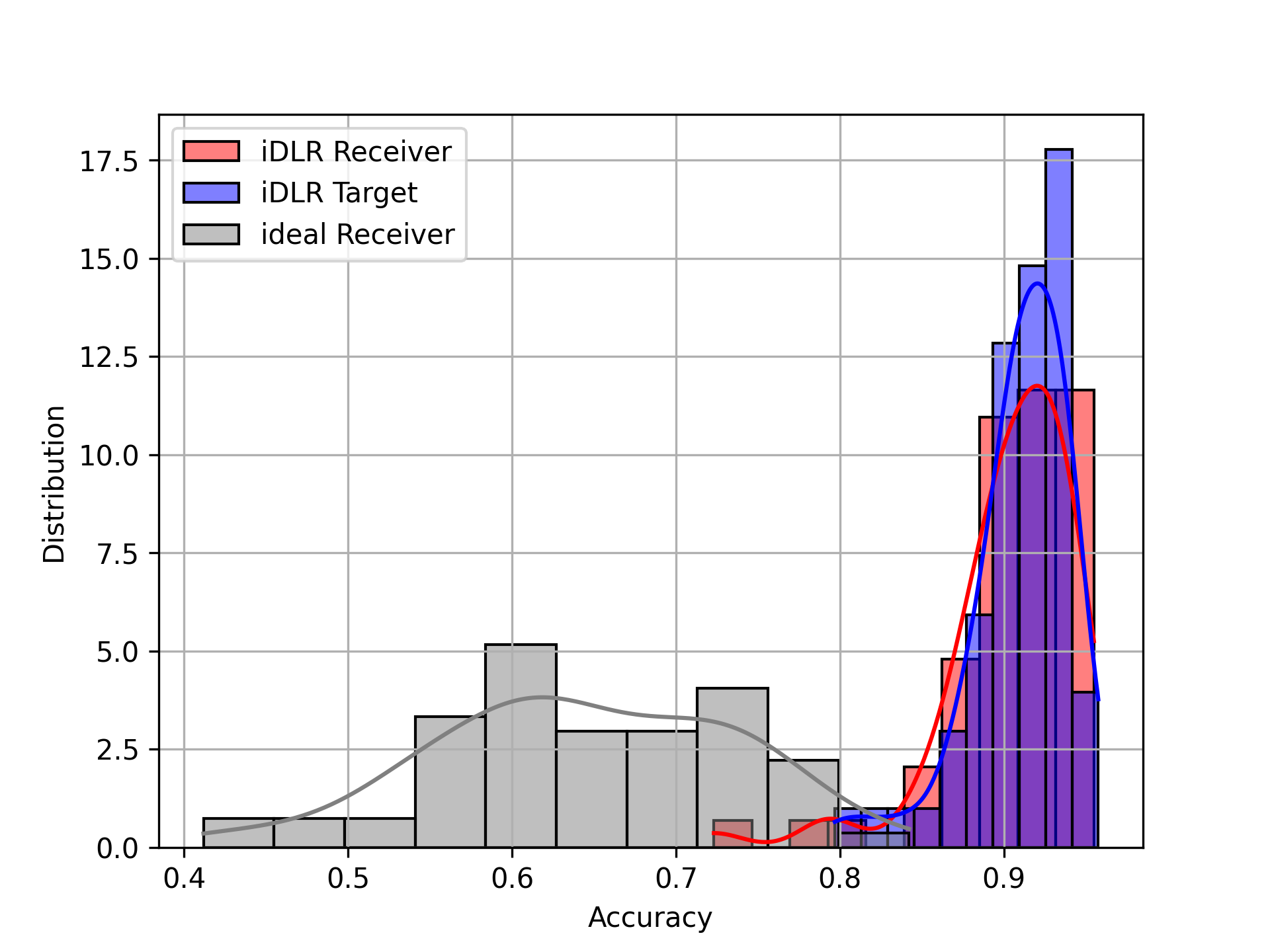} 
\caption{The distribution of the accuracy for iDLR flux density on the Lambertian Target (blue) and the receiver (red). For comparison the accuracy of the ideal heliostat assumption on the receiver is shown in gray.}
\label{fig:flux_predictions_iDLR} 
\end{figure} \noindent 
While the surface predictions are generally similar to those obtained via deflectometry, so far it is uncertain whether they are sufficiently accurate for predicting flux densities on the receiver for new sun positions. To assess this, we designed a double-extrapolation scenario—a particularly challenging case for the model. Surface predictions were generated using target images from two Lamertian Targets located at the STJ, at heights of 36~m and 43~m, primarily recorded during summer. Flux densities were then computed via raytracing on the STJ receiver, positioned 55~m above ground, curved with a radius of 4.14~m, and featuring an opening angle of 60~$\deg$, tilted 25~$\deg$ toward the ground. The simulation assumed a sun position of $(-1, -1, 1)$ in (east, north, up) coordinates—corresponding to an autumn/spring sun position in Germany. This setup represents a strong extrapolation scenario, as it significantly deviates from the sun positions under which most of the training images were acquired. For comparison, we computed the flux density predictions using iDLR-enhanced raytracing and contrasted them with predictions from deflectometry-enhanced raytracing as well as those based on the ideal heliostat assumption. Results are presented in Figure~\ref{fig:flux_predictions_iDLR}. The iDLR predictions achieve a mean accuracy of 90\% relative to deflectometry-enhanced raytracing, indicating strong extrapolation capability. However, in the case of two heliostats with relatively poor surface predictions, accuracy dropped below 80\%. This is expected, as surface inaccuracies can have a more pronounced effect in extrapolation scenarios due to underdetermined conditions. Nevertheless, for the majority of heliostats, the extrapolation performs exceptionally well, with minimal accuracy loss. By contrast, the ideal heliostat assumption achieves only 64\% accuracy, meaning that iDLR outperforms it by 26 percentage points. Additionally, the accuracy distribution of iDLR is more consistent, exhibiting a standard deviation of 4\%, compared to 9\% for the ideal heliostat assumption.
\begin{figure}[h] 
\includegraphics[width=\columnwidth]{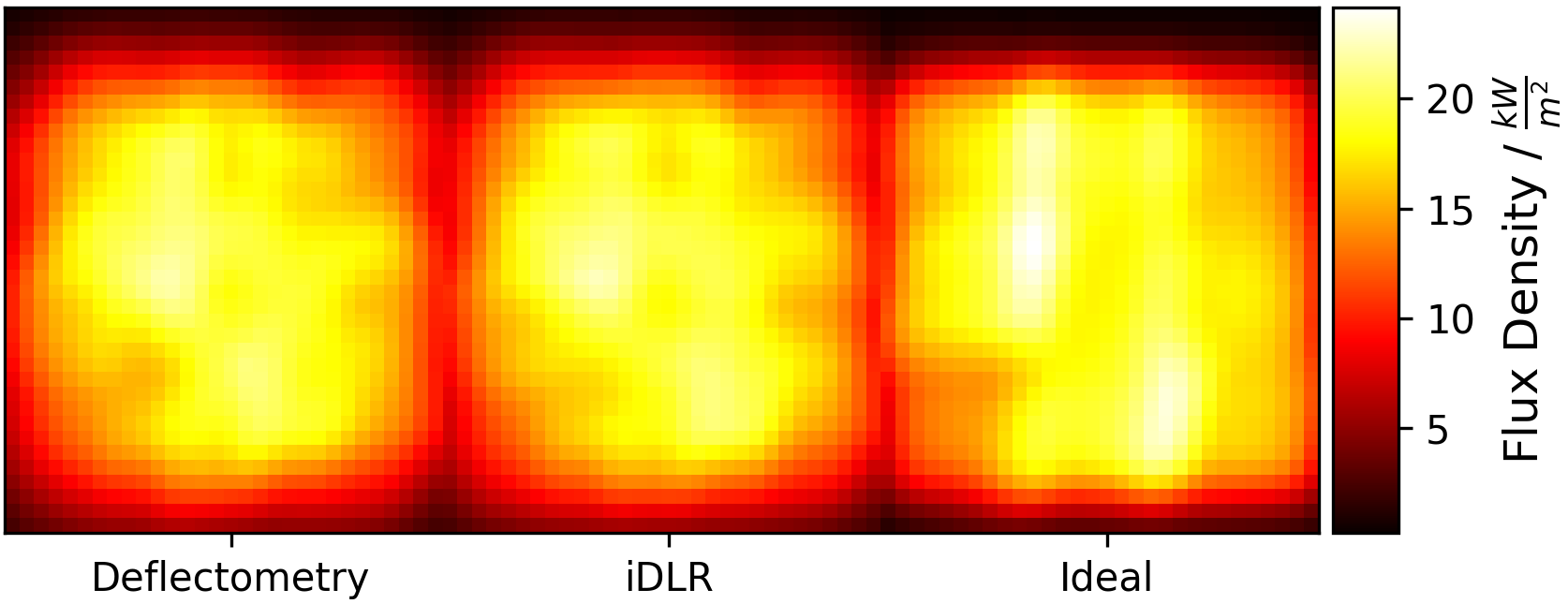} 
\caption{Accuracy of flux density predictions using iDLR enhanced raytracing compared to deflectometry enhanced raytracing. Blue lines represent predictions on a Lambertian target; red lines represent predictions on the STJ receiver in the extrapolation scenario. Gray lines indicate predictions based on an ideal heliostat assumption.} 
\label{fig:flux_predictions_iDLR} 
\end{figure}
Finally, we superposed the iDLR-based flux density predictions on the receiver and compared them to both deflectometry based predictions and the ideal heliostat assumption, as shown in Figure~\ref{fig:flux_predictions_iDLR}. The superposition was performed using six aim points arranged in a grid to distribute the irradiation on the receiver. The differences between the iDLR and deflectometry based flux density predictions on the receiver are minimal. The resulting accuracy of iDLR compared to deflectometry is 97.5\%, whereas the ideal heliostat assumption achieves only 89.5\%. 
\section{Discussion}\label{sec:Discussion} \noindent
We have combine Domain Adaptation and Domain Randomization to transfer iDLR to real data and predict heliostat surfaces using target images that were orginally taken for calibration purposes. The results show that its possible to predict heliostat surfaces with only a small loss compared to deflectometry (median MAE of 0.17~mm) for the majority of the heliostats, while there exists a risk of getting a poor surface prediction for around 16\% of the heliostats. However, as the problem is strongly underdetermined and ill-posed, we found that despite the occasionally poor surface predictions, the arising flux density prediction on the target are extremely accurate, with a median accuracy of 92\% compared to deflectometry. \newline \newline The proposed Sim-to-Real transfer approach has demonstrated strong effectiveness. Without this transfer mechanism, models trained exclusively on simulated flux densities would not generalize sufficiently to real-world target images. Nonetheless, there remains a small performance gap between simulated and real data: the median MAE on real-world data is 0.05~mm higher than on simulated data. While this discrepancy is relatively minor, it suggests potential for further refinement of the Sim-to-Real transfer pipeline to achieve even closer alignment between domains. \newline \newline 
We found that surface prediction quality improves with increasing distance from the tower. This is consistent with the simulation results reported by \citet{Lewen2025}. As discussed therein, two opposing effects contribute to this observation. First, a constant canting error leads to a reduction in underdetermination when the heliostat is positioned farther from the target. Second, surface deformations with higher spatial frequencies—such as waviness—introduce greater underdetermination for distant heliostats. As a result, canting errors become more predictable when the heliostats are located farther from the tower. While theoretically the inverse holds for waviness—i.e., the prediction of high-frequency surface features should improve at closer distances—in practice, the prediction of waviness from target images remains strongly underdetermined due to canting effects. Canting superimposes flux distribution of all four facets, making it difficult to attribute specific flux density features to individual facets. To resolve finer surface features, the model relies heavily on physical surface characteristics learned during training. Based on this learned knowledge, the model can infer finer surface deformations in a manner largely independent of the heliostat's distance from the tower. Consequently, the overall surface prediction quality increases with greater distance. However, this also implies that the effect may not generalize to other heliostat models. For example, in the case of a heliostat with non-canted facets, it is likely that the trend is reversed—i.e., prediction quality is higher for nearby heliostats compared to those farther away. \newline \newline 
Furthermore, our results show that, despite differences between the iDLR-predicted surfaces and those obtained via deflectometry, the iDLR surfaces are sufficiently accurate to enable reliable flux density predictions on the receiver, even for distant sun positions. To validate this, we designed a challenging double-extrapolation scenario: flux densities on the receiver during autumn were predicted based on surface estimates derived primarily from summer target images, acquired using a Lambertian target located around 15~m below the receiver. Despite this extrapolation in both time (seasonal variation) and space (from target to receiver/ flat surface to curved surface), the predicted flux densities retained 90\% of the accuracy achieved with deflectometry-based surfaces. This corresponds to an additional error of only 2\% due to the extrapolation, highlighting the robustness and generalization capability of iDLR surface predictions. Moreover, the superposed flux density distribution on the receiver across all 63 test and validation heliostats achieved an impressive 97.5\% agreement with deflectometry-based flux predictions on the receiver. These results strongly support the integration of iDLR-predicted surfaces into heliostat digital twins for enhanced flux density forecasting and solar field optimization. \newline \newline In comparison to manual heliostat surface measurement techniques such as deflectometry~\citep{Ulmer2011-gp}, laser scanning~\citep{photogrammetry_and_laserscanning, monterreal2017improved}, photogrammetry~\citep{Shortis1996, Pottler2005-yb, pottler2005photogrammetry, roger2010heliostat, bonanos2019heliostat}, or flux mapping~\citep{MARTINEZHERNANDEZ2023112162}, the key advantage of iDLR lies in its automation and scalability. Once trained, the iDLR framework operates at virtually zero cost and can be deployed with minimal effort—an essential requirement for integrating surface predictions into digital twins of heliostats for optimized solar field control. However, some limitations must be considered. First, iDLR requires surface measurement data for training, necessitating at least one complete measurement setup per heliostat model. While such data can be collected at any operational CSP plant, an even more convenient solution would be to perform measurements at the manufacturing site, potentially in a controlled indoor environment. Still, for heliostat models with only a limited number of units, acquiring the required \~300 surface measurements may be challenging. Despite this, in a future scenario where CSP is deployed at scale and only a few high-volume heliostat models dominate the market, several hundred surface measurements may be easy obtainable for all those heliostat models. Both manufacturers and plant operators are likely to periodically perform surface measurements for quality control, which can rapidly accumulate into sufficient training datasets. Another inherent limitation of iDLR is its statistical nature. Even though iDLR enhanced raytracing flux densities match closely with observed UNet flux densities, the underdetermined nature of the inverse problem implies that there is still uncertainty if the iDLR surface prediction is actually similar to the real heliostat surface. We found empirically that the probability of having a poor surface prediction is 16\%. This restricts the applicability of iDLR in scenarios where highly reliable surface reconstruction is critical—for instance, during acceptance testing of newly delivered heliostats—rather than for downstream tasks like flux density prediction where surface variations have hardly impact. \newline \newline 
Compared to other existing approaches that aim to derive a heliostat digital twin from target images, our method provides a significant improvement in both generalizability and physical interpretability. The post-calibration method of \citet{Zhu2022}, which only optimizes canting and focusing errors, does not model surface deformations. Our findings indicate that such deformations are critical for accurate flux density predictions, particularly at the STJ site. While the differentiable raytracer introduced by \citet{Pargmann2024} is theoretically capable of representing precise heliostat surface profiles, it suffers from the ill-posedness of the inverse problem and fails with the surface reconstruction under realistic conditions. The method proposed by \citet{Kuhl2024_flux_pred} achieves flux density predictions with accuracies around 90\% using a non-physical, purely data-driven model of the heliostat. While this is comparable to the accuracy achieved by iDLR, their predictions are restricted to the target plane, whereas iDLR makes predictions directly on the receiver. According to \citet{Kuhl2024_flux_pred}, this spatial extrapolation introduces an additional error of 2–10\%. Moreover, iDLR offers the flexibility to integrate with physical modeling frameworks: it enables simulation of soiling effects, inclusion of shading and blocking, incorporation of realistic sun shapes, and other scene-specific phenomena within ray-tracing environments. On the other hand, a clear strength of purely data-driven approaches lies in their independence from surface measurements and physical modeling—removing the need for raytracing and thereby reducing implementation complexity and required labor.
\section{Conclusion}\label{sec:conclusion_outlook} \noindent \noindent 
This work introduced the Sim-to-Real transfer of inverse Deep Learning Raytracing (iDLR) to real-world data, enabling the prediction of heliostat surface shapes from target images and supporting accurate flux density predictions under realistic conditions. Through a comprehensive evaluation across 63 heliostats, we demonstrated that iDLR surface predictions closely align with deflectometry ground truth for the vast majority of cases (84\%), achieving a median MAE of 0.17~mm. Despite occasional mispredictions—an expected consequence of the underdetermined nature of the inverse problem—the resulting flux density estimates remain highly accurate. On the Lambertian target, iDLR-enhanced raytracing achieved a median flux prediction accuracy of 92\% relative to deflectometry. Even in a challenging double-extrapolation scenario, where flux predictions on the receiver during autumn were generated from summer target images and unseen sun positions, iDLR maintained an average accuracy of 90\%, and a remarkable 97.5\% agreement when fluxes were superposed across all 63 validation and testing data. These results confirm the robustness and generalization capability of the iDLR approach. Overall, iDLR enables cost-efficient, automated, and scalable surface reconstruction, making it highly suitable for integration into digital twins of heliostats. iDLR provides a valuable middle ground between physics-based and data-driven methods, offering both interpretability and compatibility with physical simulation tools such as raytracing. These capabilities position iDLR as a key enabler for next-generation heliostat monitoring, calibration, and control.

\section*{Competing interests}\noindent
The authors declare no competing interests.

\section*{Declaration of generative AI in scientific writing}\noindent
During the preparation of this work the author(s) used large language models (LLM) (GPT4 of openAI and the LLMs by deepL) in order to improve the language. After using this tool/service, the author(s) reviewed and edited the content as needed and take(s) full responsibility for the content of the publication.
\section*{Acknowledgments}\noindent
We gratefully acknowledge the use of data from the Solar Tower Jülich, a research power plant operated by the German Aerospace Center (DLR). Additionally, we extend our thanks for the funding provided by the Helmholtz Association (HGF) for the GANCSTR project (Grant Number ZT-I-PF-5-069), under which this research was conducted. Furthermore, we thank the Helmholtz Association's Initiative and Networking Fund for funding computational resources on the HAICORE@FZJ partition.
\bibliography{lit}
\bibliographystyle{elsarticle-num-names}
\appendix
\newpage
\end{document}